\documentclass{hotnets19}
\usepackage{times}  
\usepackage{hyperref}
\usepackage{algorithm}
\usepackage{algorithmic}
\usepackage{amsmath}
\usepackage{amssymb}
\usepackage{hyperref}
\usepackage{todonotes}
\usepackage{caption}
\usepackage{subcaption}
\usepackage{tabu}
\usepackage{booktabs}
\usepackage{multirow}
\usepackage{multicol}
\usepackage{bm}
\usepackage{tabu}
\usepackage{cleveref}

\allowdisplaybreaks
\crefformat{section}{\S#2#1#3}
\Crefformat{section}{\S#2#1#3}

\newcommand{\fedavg}{\texttt{FedAvg}}
\newtheorem{conjecture}{Conjecture}
\newtheorem{definition}{Definition}

\let\oldnl\nl
\newcommand{\nonl}{\renewcommand{\nl}{\let\nl\oldnl}}

\newcommand{\tightcaption}[1]{\vspace{-0.2cm}\caption{#1}\vspace{-0.2cm}}
\newcommand{\para}[1]{\smallskip\noindent\textbf{#1}}

\hypersetup{pdfstartview=FitH,pdfpagelayout=SinglePage}

\newenvironment{packeditemize}{
\begin{list}{$\bullet$}{
\setlength{\itemsep}{1.5pt}
\setlength{\labelwidth}{8pt}
\setlength{\leftmargin}{10pt}
\setlength{\labelsep}{3pt}
\setlength{\listparindent}{\parindent}
\setlength{\parsep}{1.5pt}
\setlength{\parskip}{1.5pt}
\setlength{\topsep}{1.5pt}}}{\end{list}}

\begin{document}


\title{Enhancing the Privacy of Federated Learning\\ with Sketching}

\author{Zaoxing Liu, Tian Li, Virginia Smith, Vyas Sekar\\
Carnegie Mellon University}

\maketitle
\begin{abstract}
In response to growing concerns about user privacy, federated learning has emerged as a promising tool to train statistical models over networks of devices while keeping data localized. Federated learning methods run training tasks directly on user devices and do not share the raw user data with third parties. However, current methods still share model updates, which may contain  private information (e.g., one's weight and height), during the training process. Existing efforts that aim to improve the privacy of federated learning make compromises in one or more of the following key areas: performance (particularly communication cost), accuracy, or privacy. 
To better optimize these trade-offs, we propose that \textit{sketching algorithms} have a unique advantage in that they can provide both privacy and performance benefits while maintaining accuracy. We evaluate the feasibility of sketching-based federated learning with a prototype on three representative learning models. Our initial findings show that it is possible to provide strong privacy guarantees for federated learning without sacrificing  performance or accuracy. Our work highlights that there exists a fundamental connection between privacy and communication in distributed settings, and suggests important open problems surrounding the theoretical understanding, methodology, and system design of practical, private federated learning.
\end{abstract}
\section{Introduction}
Modern Internet-of-things devices, such as mobile phones, wearable devices, and smart homes, generate a wealth of data each day. This user data is crucial for vendors and service providers to use in order to continuously improve their products and services using machine learning. However, the processing of user data raises critical privacy concerns, which has led to the recent interest in \emph{federated learning}~\cite{mcmahan2016communication}. Federated learning explores \emph{training} statistical learning models in a distributed fashion over a massive network of user devices, while keeping the raw data on each device local in order to help preserve privacy. 

To date, although federated learning is motivated by privacy concerns, the privacy guarantees of current methods are limited. For example, the state-of-the-art method~\fedavg, proposed in~\cite{mcmahan2016communication}, only suggests that the raw user data remain local, but the individual model updates from users, such as the computed ``gradients'' in each round of the training task, are still readable by a third-party (e.g., a remote server). Communicating  model updates potentially reveals sensitive user information, such as a user's health data in a federated disease detection task (discussed in detail in~\Cref{sec:fl}). Meanwhile, current efforts that attempt to boost the privacy of federated learning build upon traditional cryptographic techniques, such as secure multi-party computation~\cite{bonawitz2017practical} and differential privacy~\cite{res_pfl,geyer2017differentially}. As federated learning is commonly deployed on low-powered, mobile user devices, existing efforts add significant communication and computation cost, making them impractical to retain user experiences.

Ideally, federated learning methods should offer \emph{privacy} by privatizing individual user updates, \emph{performance} via low communication cost, and \emph{accuracy} by providing similar convergence guarantees as the vanilla learning process. Achieving privacy while retaining accuracy and performance has been an elusive goal in machine learning~\cite{abadi2016deep,agarwal2018cpsgd,geyer2017differentially,mcmahan2017learning}, systems~\cite{bonawitz2019towards}, and theory~\cite{res_pfl,melis2015efficient} communities. 


In this work, our insight is a new connection between federated learning and sketching algorithms (sketches), a methodology that has been mainly limited in application to the areas of network measurement and databases. We find that sketches are a promising option to jointly optimize \emph{two sides of the same coin} (performance and privacy in federated learning) for two reasons: (a) First, sketches attain high accuracy with succinct data structure and have a well-studied trade-off between accuracy and memory~\cite{ams,JL-lemma}. This feature is important in attaining learning efficiency and user experiences as
massive wireless communication can cause battery draining and overheating problems from the 4G radio of mobile phones.  
(b) Second, we notice that sketches have inherent but little explored privacy benefits. In fact, most canonical sketches (e.g., Count-Min Sketch~\cite{CMSketch} and Count-Sketch~\cite{CountSketch}) do not naturally preserve data identities and require additional mechanisms to trace back~\cite{CountSketch,revsketch} to the pre-inserted identities. While this is a key  limitation of sketches in classical  measurement tasks, it  interestingly turns into a strength in the federated learning case. Moreover, recent theoretical advances~\cite{aggarwal2007privacy,melis2015efficient} have  shown that \emph{differential privacy} is achievable on sketches with additional modifications. 
These properties make sketches an attractive alternative to enhance federated learning meet the privacy, performance, and accuracy goals. 


\begin{figure}[t]
    \centering
    \includegraphics[width=0.49\textwidth]{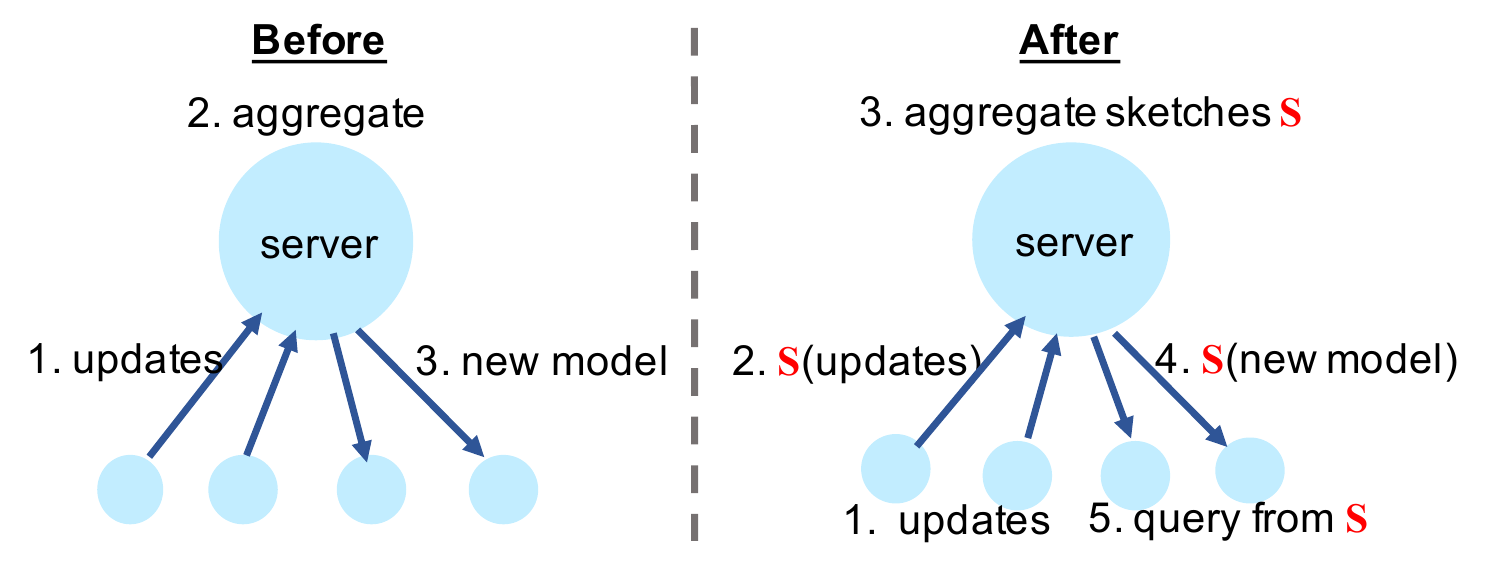}
    \vspace{-2mm}
    \tightcaption{Conceptual design of federated learning with sketching (S).}
    \label{fig:concept}
\end{figure}

Our vision in this work is to bring to light these sketching techniques to practical federated learning systems and show a viable path towards a \emph{private federated learning} system. The key challenge of realizing such a vision is in accounting for federated learning's unique distributed workflow and designing appropriate private  communication mechanisms. Specifically, we consider the following questions: (1) How can sketches be used in federated learning frameworks to boost the overall privacy? (2) What current sketches are useful in terms of privacy and accuracy? (3) Can we further improve the privacy guarantees of current sketches? 

As a concrete start to answer these questions, we provide a preliminary design that incurs only small changes to current federated learning frameworks, as shown in Figure~\ref{fig:concept}.  Our approach leverages sketching on the updates sent between users and the central server to prevent the identities of private user information from being revealed.
To verify the feasibility of using sketches to privatize individual model updates without significant accuracy impact, we implement a proof-of-concept federated learning simulation with Count Sketch~\cite{CountSketch} to train three representative models: Linear Regression, Multi-layer Perceptron (MLP), and Recurrent Neural Networks (RNN). Our simulation demonstrates that extra privacy can be added into federated learning for ``free''  as sketches privatize the original data and save the communication cost by 10$\times$ with small  errors. 

Our vision, if successful, has the ability to dramatically mitigate the concerns of user privacy in federated learning with  strengthened confidence in user experiences. We also identify  further directions to explore the theoretical understanding of the privacy features of sketches, along with system and privacy requirements from various federated learning tasks, and design appropriate sketching-based federated learning framework based on user needs in~\Cref{sec:discussion}.




\section{Background}
\subsection{Federated Learning}\label{sec:fl}
Federated learning algorithms assign learning jobs to hundreds to millions of remote devices with their self-generated data. 
Instead of sending the local data  to the cloud to jointly learn a model, in federated settings user data and training remain local on each device. This design benefits user privacy and reduces the communication of the training process~\cite{li2019federated}.

\para{Learning scenarios and their privacy issues.} Federated learning is useful for many applications. Unfortunately, as we show in the following representative scenarios, user privacy can still be an issue even when user data remains local. 

\begin{figure}
    \centering
    \includegraphics[width=0.45\textwidth]{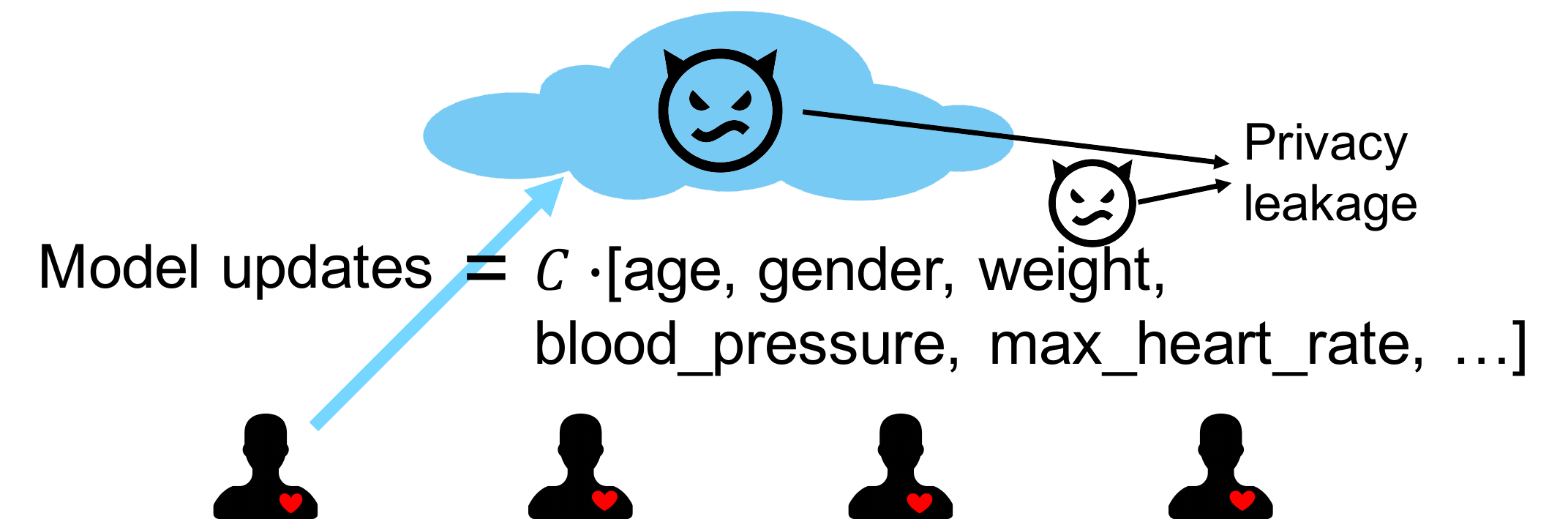}
    \vspace{2mm}
    \tightcaption{Privacy leakage in S1: Inferring user health data from individual model updates.}
    \label{fig:example_evil}
\end{figure}

{\bf \underline{S1:}} \emph{Heart attack risk detection from wearable devices.} Consider training a linear model via minimizing the empirical loss over historical health
data distributed across wearable devices. The loss function of such a linear model can be defined as $F(w; x, y)=(w \cdot x-y)^2$ where $F$ means the gap between the model's predictions and the ground-truth label, $w$ denotes the learnable parameters, and $x$ represents user data (i.e., related health information). 
As shown in Figure \ref{fig:example_evil}, the gradient of $F(w; x,y)$, used by the de facto optimizers (min-batch) Stochastic Gradient Descent (SGD) in machine learning training, is $c \cdot x$ --- some constant $c$ multiplied by the raw data $x$. Hence, sending gradients (or model updates) may be equivalent to sending raw data.

{\bf \underline{S2:}} \emph{Next word prediction on mobile phones.} In order to enhance user experiences during typing on the phones, some applications such as the Google keyboard (Gboard) leverage federated learning to jointly learn a next-word prediction model over users' historical text data and deploy the model on the virtual keyboard~\cite{hard2018federated}. In this scenario, the Recurrent Neural Networks (RNNs), a state-of-the-art neural network model is used to handle sequence data such as texts. RNNs are known to have the ability to memorize and expose sensitive, unique patterns in the training data~\cite{carlini2019secret}. For instance, if someone's credit numbers are typed before, the resulting model may contain the credit card numbers in some way. As shown in~\cite{carlini2019secret}, one can efficiently extract the secret sequences from the final model. 

{\bf \underline{S3:}} \emph{Face recognition on cameras.} Another application to consider is performing distributed face recognition on a calibrated camera sensor network~\cite{gaynor2015distributed}. In such settings, an individual's private facial information is vulnerable to several specific attacks, and it is possible for attackers to reconstruct face images by accessing to the learned model~\cite{fredrikson2015model}. A basic idea of such attacks is to enumerate different combinations of the target's face pixels, and identify the image such that the model prediction on that image has the maximum confidence score on the target's category. 

\subsection{Private Federated Learning Vision}\label{bg:vision}
Whereas traditional federated learning is concerned mainly with achieving a reasonable accuracy on some machine learning task, our vision is to achieve privacy, performance, and accuracy simultaneously. A first key step towards this goal is providing a concrete definition of privacy in the federated setting. Indeed, in related work on privacy and federated learning, we observe a variety of definitions regarding the privacy guarantees~\cite{ abadi2016deep,bonawitz2017practical,mcmahan2017learning}.
To help navigate this space, we introduce a taxonomy of commonly-used notions of privacy that are applicable to federated learning below: 
\begin{packeditemize}
    \item \emph{Level-0:~No privacy}. Raw user information (before local model computations) is shared with other parties. 
    \item \emph{Level-1:~Raw data protection}. Raw user information is not shared, but the computed individual model updates (e.g., individual gradient data) are revealed to other parties. Traditional federated learning (e.g., \cite{mcmahan2016communication}) reaches this level as it does not share the user raw data by offloading the training process to user devices. 
    \item \emph{Level-2:~Global privacy}. Raw user information is not shared and the individual model updates are private to other untrusted third parties other than the aggregation server, e.g., user updates are differentially private in a global manner~\cite{geyer2017differentially,mcmahan2017learning}. In this level, the aggregation server (e.g., Google or Apple) is considered a trusted party to see the individual updates.
    \item \emph{Level-3:~Local privacy}. Raw user information is not shared and individual model updates are private to all third parties (including the server) with different privacy guarantees~\cite{abadi2016deep}. For instance, the local model updates can be differentially  private or total ciphertext to any third parties.
    \item \emph{Level-4:~Holy grail}. Any information about the user, model updates and the final trained model are private to any third parties (ad omnia privacy!). Every participant only has access to the predictions from a trained black-box model, which does not reveal any potential private information in the final trained model.
\end{packeditemize}

While different proposals are trying to achieve different levels of privacy for the learning process, the performance and accuracy are also essentially important. We define performance as no added communication cost or reduced communication compared with vanilla federated learning algorithms. Further, the accuracy is defined as the accuracy of each round's model update and the rate of convergence. The convergence in the learning tasks means that the trained model generally achieves a converged prediction accuracy after a number of rounds in  training.

\para{Existing efforts on improving the privacy of federated learning.} We are not the first to point out the privacy issue of federated learning~\cite{res_pfl,bonawitz2017practical,geyer2017differentially,mcmahan2017learning}. For instance, \cite{bonawitz2017practical} brings secure multi-party computation (SMC) into federated learning. Traditionally, SMC achieves a very high privacy guarantee by allowing multiple parties to jointly compute a function without revealing the input from each party. Unsurprisingly, SMC has complex cryptographic mechanisms~\cite{smc}.
To gain the feasibility of SMC on federated learning, the authors introduce a relaxed version of SMC that retains level-3 privacy. However, their relaxed SMC still incurs significant extra communication cost (up to $5\times$). 

The authors of \cite{mcmahan2017learning, geyer2017differentially} bring differential privacy~\cite{dp_original} in training neural networks and offer level-2 privacy, as they assume the server is trusted to see the model updates. However, the proposed mechanism is impractical as it needs careful choices on a number of hyper-parameters with impacted communication and accuracy.

In summary, existing work achieves level-2 to level-3 privacy while sacrificing performance (e.g., relaxed SMC with many round trips of added communications~\cite{bonawitz2017practical}) and accuracy (e.g., adding Gaussian noise to individual user updates~\cite{res_pfl}). Our vision is to achieve (level-3 or higher) privacy for federated learning tasks, while maintaining or improving the accuracy and performance of traditional federated learning frameworks. 

\begin{figure*}[t]
    \centering
    \includegraphics[width=0.94\textwidth]{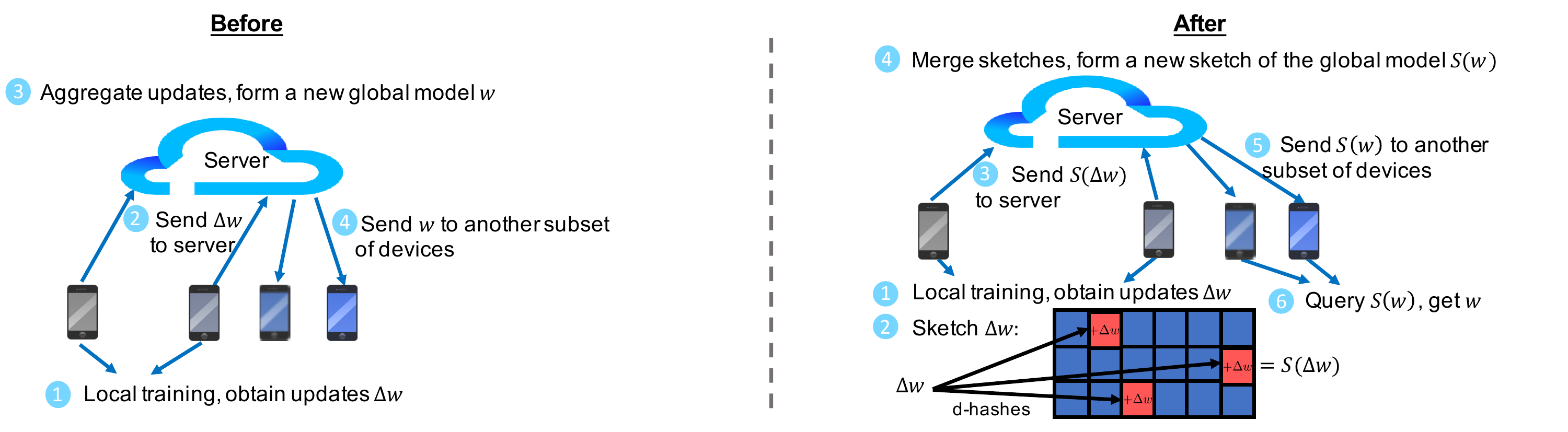}
    \vspace{-2mm}
    \caption{An overview of the learning pipeline before and after applying sketches}
    \label{fig:overview}
\end{figure*}

\begin{figure}[t]
\centering
\includegraphics[width=0.9\linewidth]{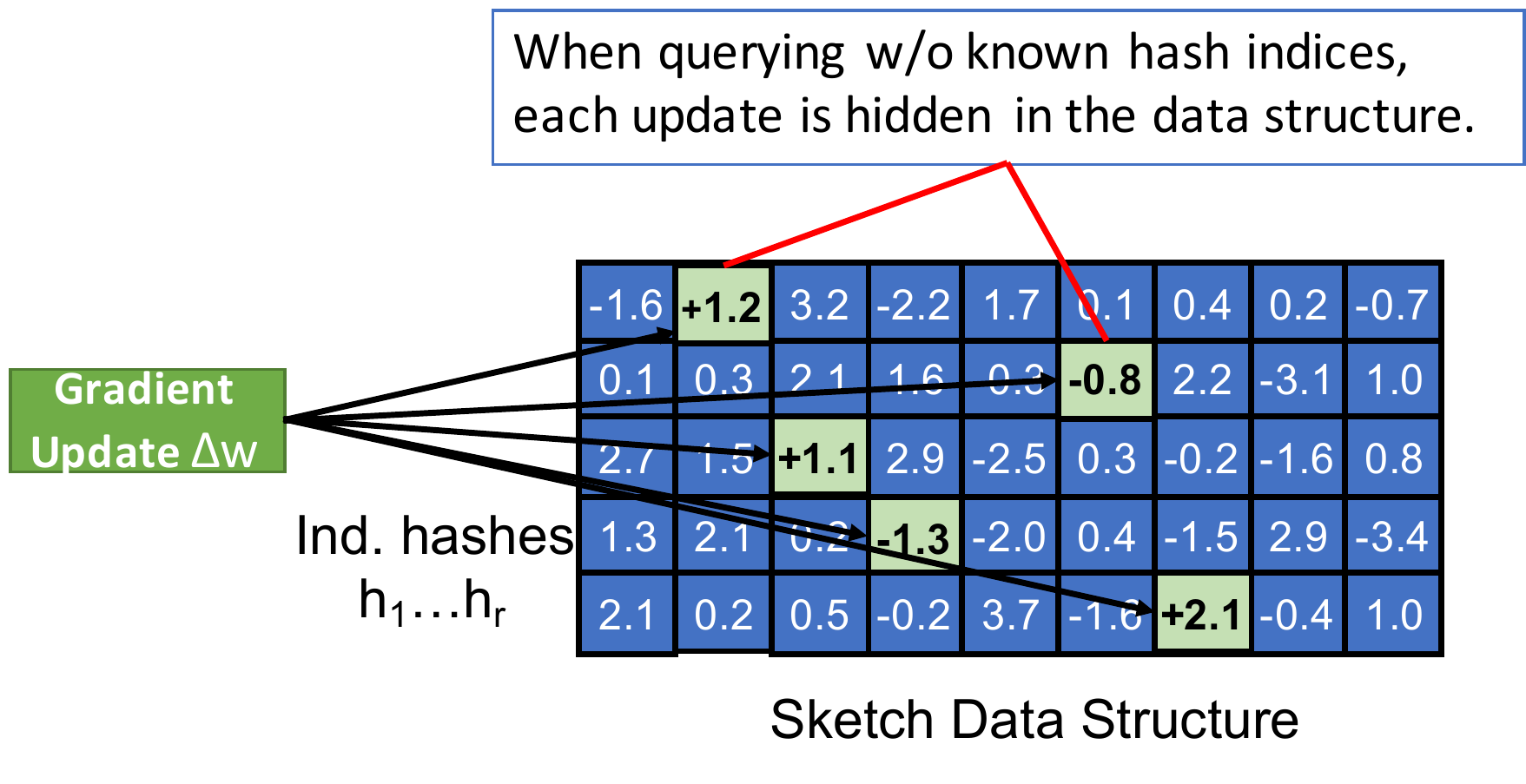}
\vspace{-2mm}
\tightcaption{Sketch data structure and its privacy property.}
\label{fig:sketch}
\end{figure}

\subsection{Privacy Features of Sketches}\label{bg:privay}

To fulfill our vision of private federated learning, we argue that sketches are a promising primitive for federated learning based on their privacy benefits. In this section, we discuss the existing efforts on private sketches and then provide our own conjectures about sketch privacy to be proven as future work.

\para{Differential privacy.} Intuitively, differential privacy guarantees a party to share a data structure to  make  sure  that  only  a  bounded  amount  of  information  is  leaked. We give the formal definition of differential privacy below:
\begin{definition}[$\epsilon$-differential privacy~\cite{dp_original}]
A \\randomized mechanism $M$ satisfies $\epsilon$-differential privacy, for all data sets $D_1$ and $D_2$ differing by up to one element,  and for any output $S$ of $M$,
\begin{equation*}
    Pr[M(D_1) \in S] \le e^{\epsilon} \cdot Pr[M(D_2) \in S]
\end{equation*}
\end{definition}


We observe that recent work from Melis et al.~\cite{melis2015efficient} explores a differential privacy guarantee derived from modified Count-Min Sketch and Count-Sketch, and use them to build a private recommendation system. Their approach leverages standard Laplace mechanism~\cite{laplace} to add noise into the sketch data structures in order to gain $\epsilon$-differential privacy. This sketch-based differential privacy requires less added noise than adding noise directly into model updates.

Although \cite{melis2015efficient} shows achievable differential privacy for sketches, the privacy benefits of  vanilla sketches have not been explored. It is unclear what privacy features sketches
can offer and what trade-off between privacy and accuracy sketches can provide, under different assumptions. One assumption about federated learning is that the scale of user devices and gradient updates is significantly larger than the traditional distributed learning. Thus, based on our observations on the theoretical analysis of sketches, we make two conjectures about the privacy guarantees of sketches for further exploration.

\begin{conjecture}\label{c1}
Canonical sketches (e.g., Count Sketch and Count-Min) can potentially achieve $\epsilon m$-differential privacy when the model space is significantly larger (in terms of some polynomial in $m$) than the size of the sketch.
\end{conjecture}

Intuitively, if the number of inserted items is sufficiently large, a few insertions or removals from the sketch will not incur noticeable change to the counters in the data structure. If Conjecture~\ref{c1} can be proven with quantified $m$, when model space is large enough, we can leverage sketches to achieve reasonable level-3 differential privacy without added noise, and achieve ideal performance and accuracy for the training process. When model space is not large, we have the following Conjecture~\ref{c2} to bound the privacy leakage.

\begin{conjecture}\label{c2}
When model space is not large enough (in some polynomial of $m$) compared to the sketch size, sketches lose differential privacy, but still prevent the reconstruction of original identities with high probability.
\end{conjecture}

Conjecture~\ref{c2} is intuitively true as depicted in Figure~\ref{fig:sketch} as a third party cannot retrieve the original identities better than random guessing. A key requirement to achieve this conjecture may be the use of cryptographic hash functions, such as SHA-256, which offers strong (mathematical) irreversibility.

\section{Overview}\label{sec:overview}
In this section, we overview our basic design by answering the following three design questions. Our design shows a viable path towards realizing our private federated learning vision.

\para{Question 1: How can sketches be used in a federated learning setting?} In federated learning as shown in the left part of Figure~\ref{fig:overview}, on each training round, the server requests to a (random) subset of devices for their local updates. Depending on the size of the machine learning model, the computed local updates can be a vector of up to millions of numbers. These update numbers are needed to be sent to the server for aggregation and the merged updates will be distributed to another subset of devices for further training. This training process is presented in~Algorithm~\ref{alg:fedavg:vanilla}. In this case, the vanilla update data can be read by any third parties along the path and is transmitted over the network for at least one (large and slow) round trip time. Since we observe that sketches obfuscate the original data using independent hash functions and thus have privacy features, we can leverage them to conceal the identities of each round of model updates with user-owned secret hash indices and seeds.

At first glance it is not clear that  ``sketching'' the updates will  preserve the accuracy in practice.  However, we notice that, among the vector of numbers in each device update, only a portion of most ``significant'' coordinates are used to train a model as the computed coordinates are usually skewed and follow a nonuniform distribution~\cite{jiang2018sketchml}. Given this fact, as depicted in the right part of~\Cref{fig:sketch}, we place sketches into the computation of local updates to compress the user updates from up to millions of numbers to tens of thousands. The pseudocode of this workflow is presented in Algorithm~\ref{alg:fedavg:sketch}. Based on the accuracy vs. space trade-off of sketches, we can pick a reasonable compression ratio (e.g., 10$\times$) for communication cost reduction while maintaining guaranteed fidelity.

\para{Question 2: What current sketches are useful in terms of privacy and accuracy?} Sketches are commonly used for applications in network measurement and databases. For instance, we have
Count-Min Sketch~\cite{CMSketch} to track the heavy hitters with $L_1$\footnote{$L_1\triangleq\sum f_x$ refers to the first norm of the flow frequency vector of the workload.} guarantee, Count Sketch~\cite{CountSketch} for $L_2$ heavy hitters, and other sketches for other metrics such as entropy estimation~\cite{Chakrabarti_entropy,Entropy1,univmon}, second frequency moment~\cite{ams,univmon}. It is natural to ask what sketch(s) will be particularly useful for various federated learning tasks. We expect that the actual choice of sketches should depend on the distribution of the data set, e.g., if the actual gradient coordinates are more uniformly distributed than other more skewed distributions, we can choose Count Sketch over Count-Min for a potentially better accuracy as Count-Min always overestimates the actual numbers and Count Sketch removes this estimation bias. We demonstrate the accuracy of Count Sketch in the evaluation (\Cref{sec:eval}).

\para{Question 3: How to achieve higher-level privacy with improvements over current sketches?} As we described in~\cref{bg:privay}, original sketches can potentially guarantee reconstruction privacy and achieve differential privacy. However, such guarantees are not the highest privacy feature we can think of. To achieve higher privacy with sketches, we need to consider various federated learning scenarios and build additional system and theory hammers. We leave new theoretical and systematic approaches as open challenges in the discussion~(\Cref{sec:discussion}).

\setlength{\textfloatsep}{2pt}
\begin{algorithm}[t]
        \begin{algorithmic}[1]
	        \FOR  {$t=0, \cdots, \text{round}-1$}
		        \STATE  Server samples a subset of $K$ devices  (each device is chosen with prob. proportional to the number of local data points)
		        \STATE Server sends the current model $w^t$ to chosen devices
		        \STATE Each device $k$ performs local training to obtain $w_k^{t+1}$
		        \STATE Each chosen device $k$ sends $w_k^{t+1}$ back to the server
		        \STATE Server aggregates the $w$'s as {\small$w^{t+1} = \frac{1}{K}\sum_{k} w_k^{t+1}$}
	    \ENDFOR
	  \end{algorithmic}
	  \caption{Federated learning algorithm (\fedavg).}\label{alg:fedavg:vanilla}
\end{algorithm}

\section{Preliminary Evaluation}
\label{sec:eval}

\begin{figure*}[t]
    \centering
    \begin{subfigure}{0.33\textwidth}
    \includegraphics[width=\textwidth]{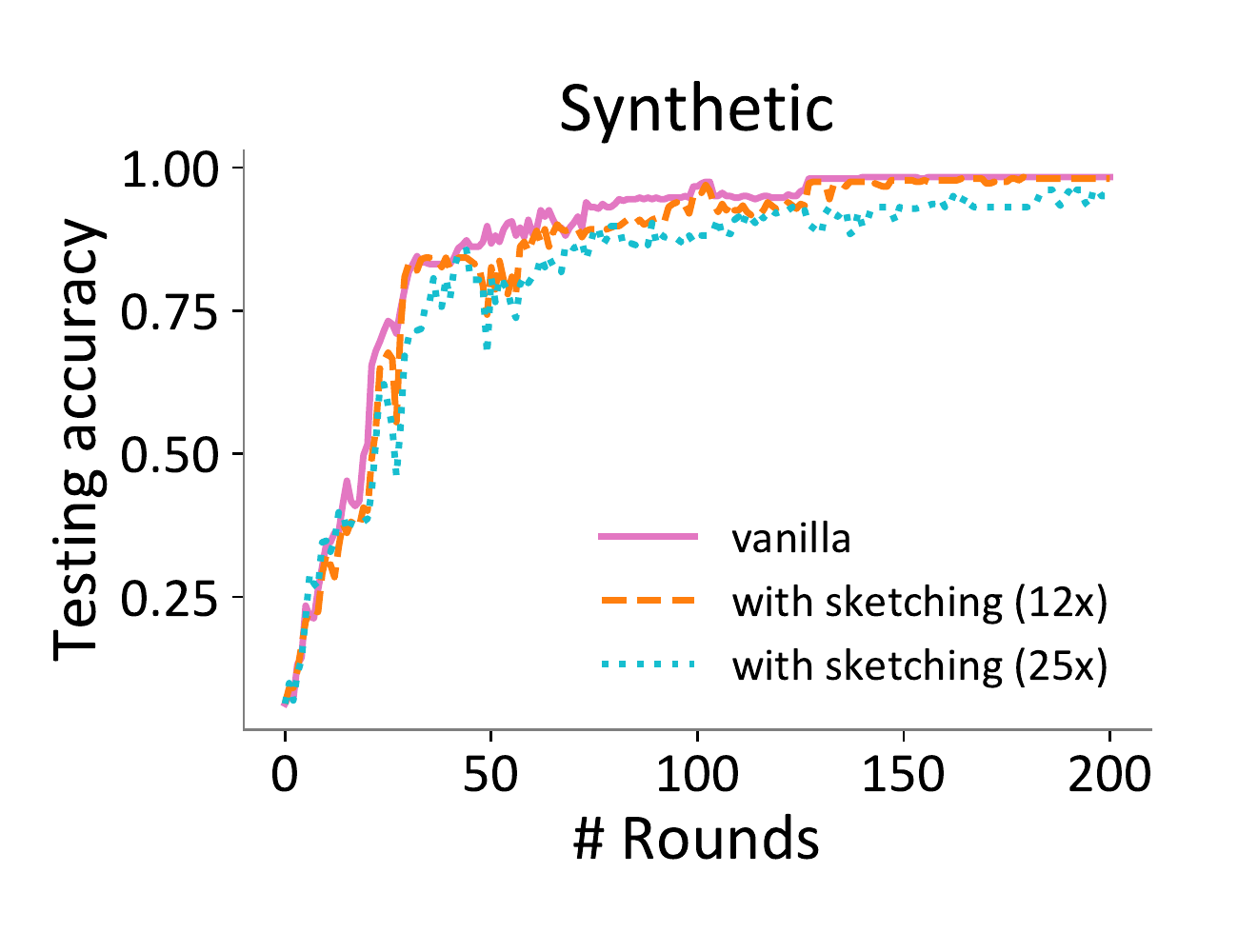}
    \end{subfigure}
    \hfill
    \begin{subfigure}{0.33\textwidth}
        \includegraphics[width=\textwidth]{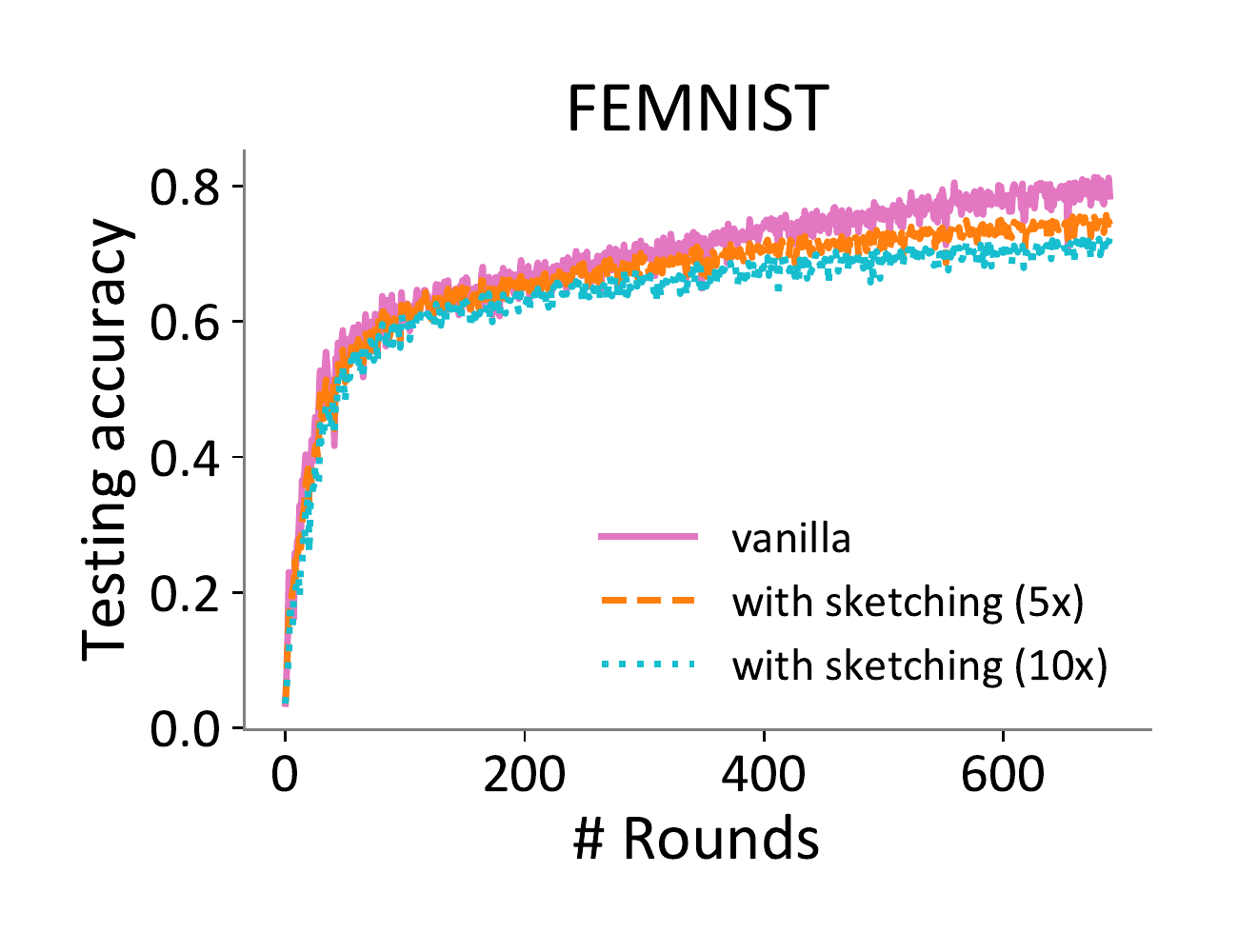}
    \end{subfigure}
    \hfill
    \begin{subfigure}{0.33\textwidth}
    \includegraphics[width=\textwidth]{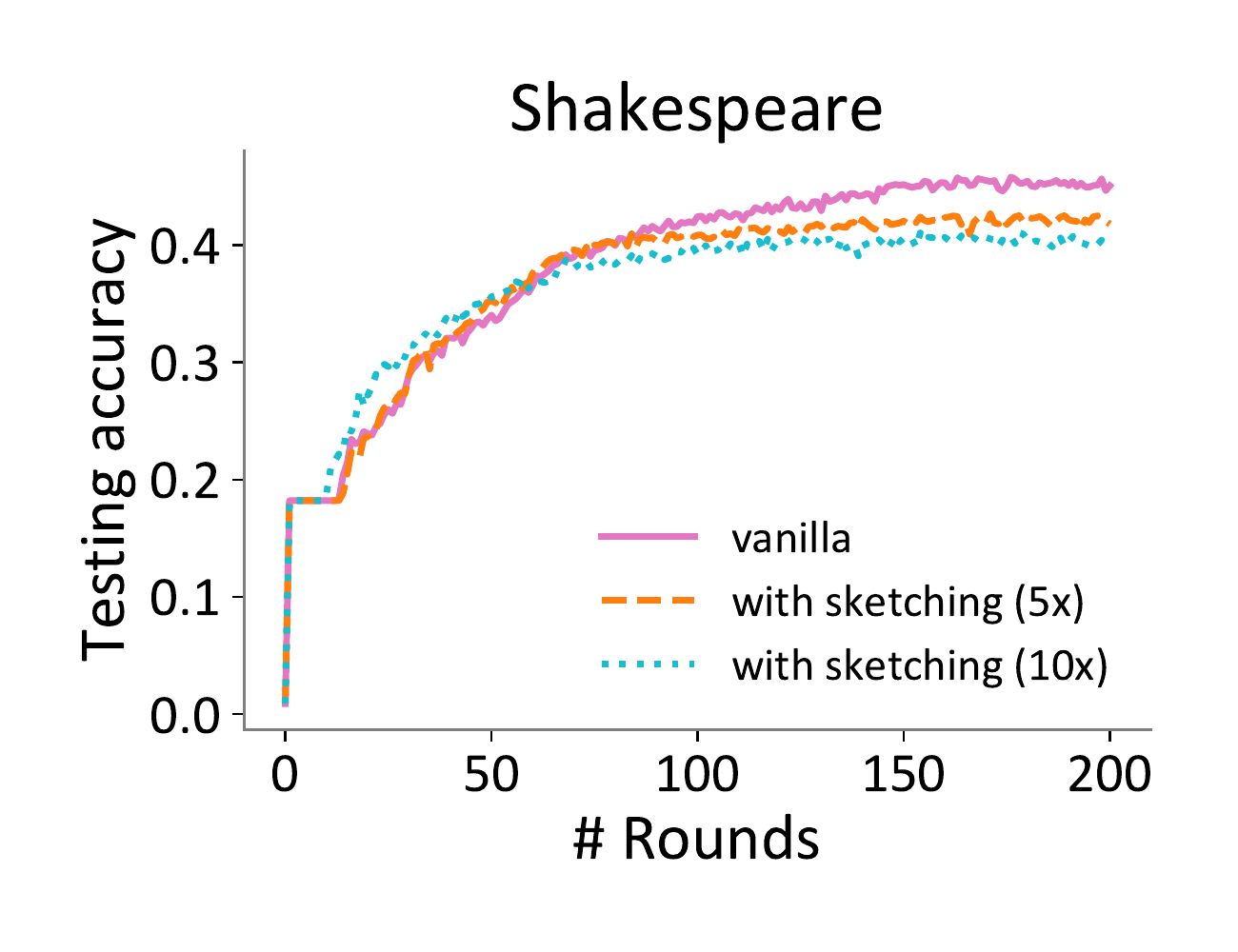}
    \end{subfigure}
    \begin{subfigure}{0.33\textwidth}
    \includegraphics[width=\textwidth]{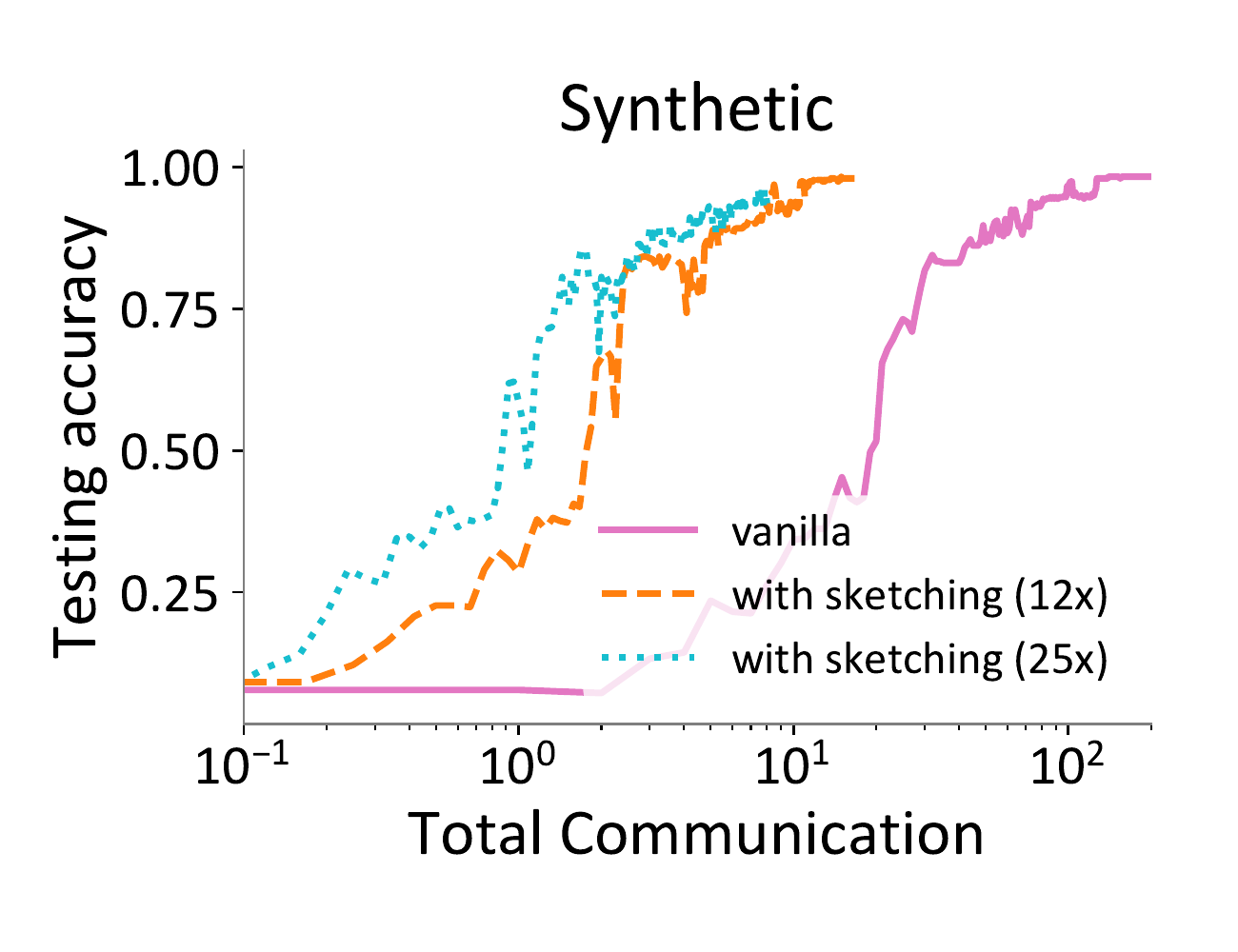}
    \end{subfigure}
    \hfill
    \begin{subfigure}{0.33\textwidth}
        \includegraphics[width=\textwidth]{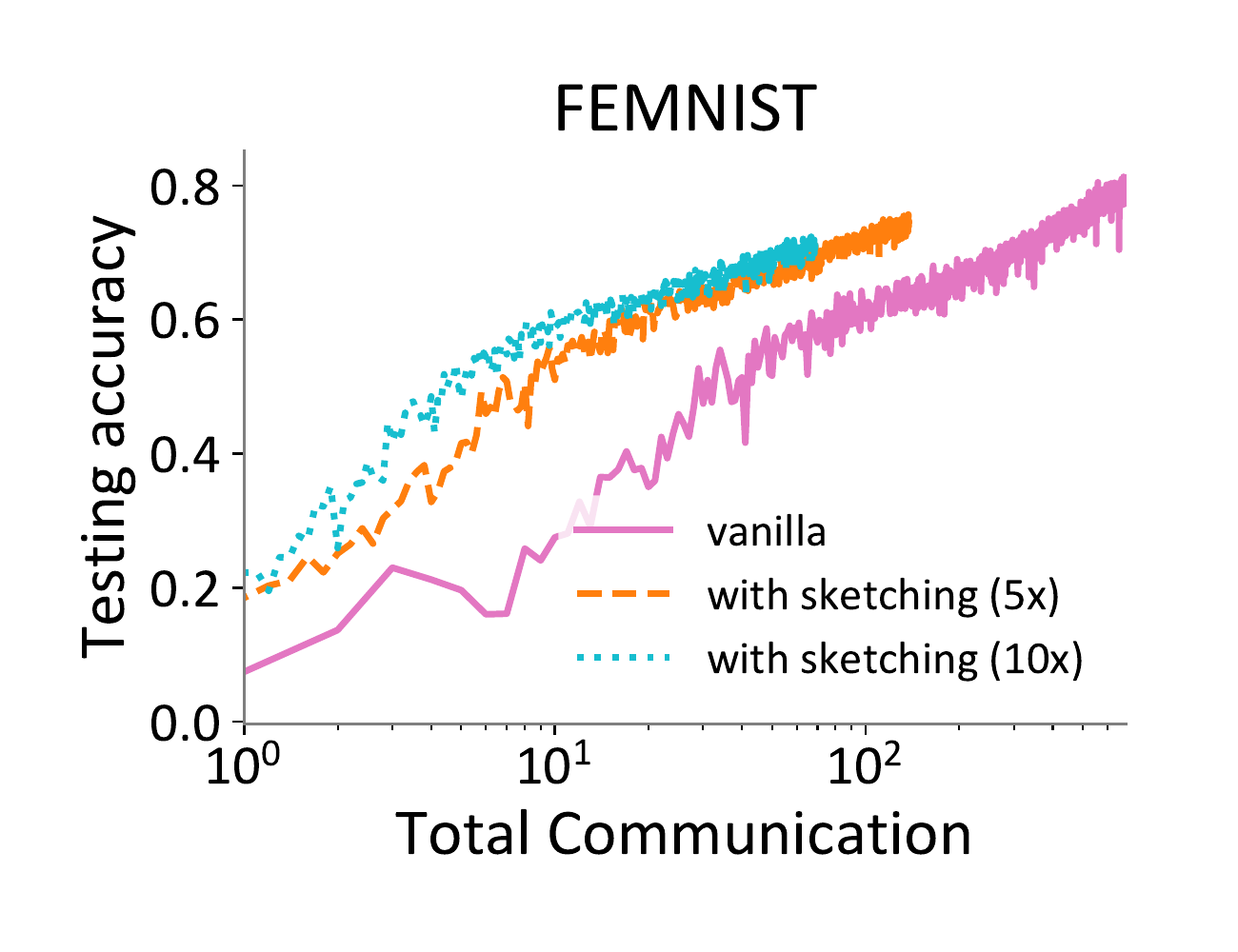}
    \end{subfigure}
    \hfill
    \begin{subfigure}{0.33\textwidth}
    \includegraphics[width=\textwidth]{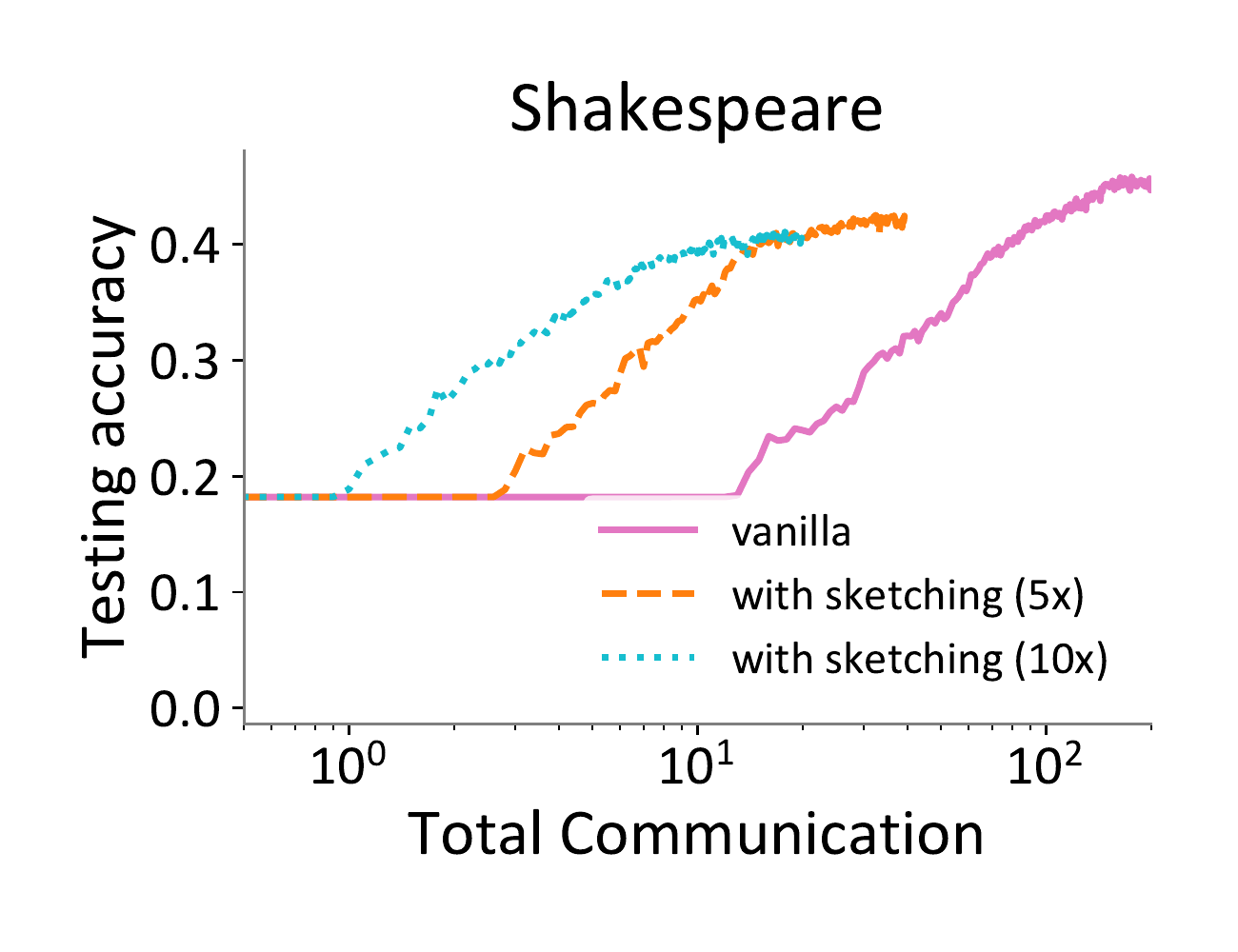}
    \end{subfigure}
    \vspace{-2mm}
    \tightcaption{The convergence on the Synthetic (Linear Regression), FEMNIST (MLP), and Shakespeare (RNN) datasets. Applying the basic Count Sketch is able to achieve significant convergence speed up with an up to $25\times$  compression ratio. We note here that the total amount of communication is in log scale.}
    \label{fig:restuls}
\end{figure*}

In this section, we  evaluate the feasibility of sketches in federated learning tasks, and demonstrate that sketches compress and privatize the communication in the federated setting with small added errors. 


\subsection{Methodology}

\para{Datasets and models.}
Federated datasets are generated by local devices in a non-identical distributed fashion. We evaluate using three different datasets and models curated from previous work benchmarking federated settings~\cite{caldas2018leaf} and recent work in federated learning: 
(a) Synthetic dataset similar to that in~\cite{shamir2014communication}. We tweak it to be highly heterogeneous (both in terms of data distribution and the number of data points) across devices. The model is linear regression classification. 
(b) Federated Extended MNIST  (FEMNIST)~\cite{cohen2017emnist} image dataset. We subsample the 26 lower-case characters for image classification, and assign 19 classes to each device in order to simulate the heterogeneity distribution. We apply a Multi-layer Perceptron (2 layers) model to this dataset. (c) Text data built from \emph{The Complete Works of William Shakespeare}~\cite{mcmahan2016communication} (Shakespeare), where each device corresponds to a speaking role in the play. We use a recurrent neural network containing two Long Short Term Memory (LSTM) layers to train a next-word prediction model. The statistics of three datasets and the number of model parameters are summarized in Table~\ref{table:data}.

\begin{algorithm}[t]
        \begin{algorithmic}[1]
	        \FOR  {$t=0, \cdots, \text{round}-1$}
		        \STATE Server samples a subset of $K$ devices  (each device is chosen with prob. proportional to the number of local data points)
		        \IF {$t>0$}
		        \STATE Server sends the sketched global model $\bm{S}(\Delta w^t)$ to all chosen devices
		        \STATE Each chosen device queries $\bm{S}(\Delta w^t)$ for $\approx\Delta w^{t}$
		        \STATE Each device $k$: $w_k^t = w_k^{t-1} + \Delta w^t$
		        \ENDIF
		        \STATE Each device $k$ performs local training: $\Delta w_k^{t+1}$
		        \STATE Each device $k$ sketches the updates: $\bm{S}(\Delta w_k^{t+1})$
		        \STATE Each device $k$ sends $\bm{S}(\Delta w_k^{t+1})$ back to the server
		        \STATE Server aggregates: { $\bm{S}(\Delta w^{t+1}) = \frac{1}{K}\sum_{k} \bm{S}(\Delta w_k^{t+1})$}
	    \ENDFOR
	  \end{algorithmic}
	  \caption{Federated learning algorithm with sketching.}\label{alg:fedavg:sketch}
\end{algorithm}

\begin{table}[t]
	\centering
	\small
	\begin{tabular}{ l | c | c | c } 
			\toprule
			\textbf{Dataset} & \textbf{Devices} & \textbf{Samples/device} & \textbf{Model params.} \\
			& & mean (stdev) & \\
			\hline
			Synthetic & 30 & 115 (58) &  6,010  \\
			FEMNIST & 200 & 94 (91)  & 51,930 \\
			Shakespeare & 46 & 742 (578) &  36,720 \\
			\bottomrule
	\end{tabular}
	\caption{Statistics of Federated Datasets and Models}
	\label{table:data}
\end{table}

\para{Implementation.} We simulate a federated learning setup (one server and $N$ devices) on one machine with 2 Intel$^\text{\textregistered}$ Xeon$^\text{\textregistered}$ E5-2650 v4 CPUs and 8 Nvidia$^\text{\textregistered}$ 1080Ti GPUs. We implement the original federated learning algorithm proposed in~\cite{mcmahan2016communication} and apply Count Sketch to preserve the most significant weights in the model updates (as demonstrated in Figure \ref{fig:overview}). To recover the model updates from the sketch, we query the median of independent sketched numbers. For the other two datasets with neural network models, we query the median of top-20\% model updates in order to boost the convergence. 
All code is implemented in Tensorflow v1.10.1~\cite{abadi2016tensorflow}.

\para{Parameters.} We fix the number of counter arrays in sketches to be 5 for all datasets and explore the effects of varying the number of counters (corresponds to different compression ratios). For each dataset, we tune all hyper-parameters such as learning rate and batch size on the baseline \fedavg~ algorithm without user privacy guarantees, and use the same parameters for all experiments on that dataset.

\para{Protocols.}  We plot the testing accuracy versus communication rounds. 
In practice, sketching will reduce the amount of communication significantly, while introducing modest local computation cost. If measuring accuracy versus real training time (a combination of computation time and communication time), we expect that our approach with sketches will perform even better.

\subsection{Results}\label{sec:results}

As shown in Figure \ref{fig:restuls}, our learning simulation with Count Sketch is able to achieve up to 10$\times$ compression ratio with small accuracy loss (on average). We fix the number of hash functions and vary the number of counters in the sketch to achieve different compression ratios. As expected, with higher compression ratio, sketched training process leads to a lower accuracy, given the trade-off between accuracy and memory in Count Sketch.
On the FEMNIST and Shakespeare datasets with non-linear models, we observe that the testing accuracy tends to be slightly worse than the vanilla algorithm particularly when the algorithm is close to convergence. This is due to the fact that model updates would converge to near zero at the end. Our static sketch may not recover such small values with high accuracy. To further improve this accuracy, we should consider adaptive sketching as a future work described in the next section.

\para{Privacy Discussions.}  We now provide an intuitive explanation on the privacy of our approach. 
Define the model update from device $k$ as an $n$-dimensional vector $\Delta w_k$. Assuming that the server knows which sketch table $\bm{S}(\Delta w_k)$ belongs to device $k$, even if the server could recover each element in $\Delta w_k$ from $\bm{S}(\Delta w_k)$ with 100\% accuracy (which is impossible), it is difficult for the server to know the identity of each estimated element in the vector. Therefore, the probability of recovering the value on the $i$-th index $\Delta w_k[i]$ would be $\frac{1}{n}$. In practice, the probability of estimating the real value $\Delta w_k[i]$ (for any $i$) should be less than $\frac{1}{n}$. 


\section{Discussion}\label{sec:discussion}
We conclude by highlighting a subset of new and exciting challenges that this work opens up.

\para{Theoretical understanding and enhancement of the privacy guarantees from sketches.} In~\Cref{bg:privay}, we briefly overview the potential privacy features of sketches with existing work, and come up with two conjectures about the privacy of vanilla sketches. While attempting to prove/disprove the conjectures, it is natural to gain a more comprehensive theoretical understanding of what privacy benefits sketches can possibly provide with or without additional mechanisms, e.g., if unmodified sketches cannot achieve differential privacy, what information is leaked and can we bound this leakage using probabilistic tools~\cite{prob_method}? 

To strengthen the privacy feature of sketches, as current sketches are designed mainly for network and database scenarios without explicit privacy considerations, we can consider tailoring sketches with tools such as Gaussian or Laplace noise~\cite{laplace}, oblivious transfer~\cite{oblivous_transfer}, and homomorphic encryption~\cite{homomorphic}. In this direction, our latest efforts~\cite{diffsketch} prove that we can leverage and improve Count Sketch to achieve high differential privacy on distributed SDG and federated learning scenarios.

\para{Adaptive sketching for various federated learning scenarios.} In diverse federated learning workloads, the device data and model update distributions vary by user and time. To approach optimal learning accuracy and communication efficiency, it is unlikely that one specific sketch can fit them all. Naturally, we need to devise an adaptive sketching mechanism based on different model update distributions. When the distribution changes statistically, we might need to adjust the sketches accordingly with different types of sketching algorithms, different compression ratios, and different hash functions. Achieving this requires a full theoretical understanding of sketching's privacy and accuracy guarantees. 

\para{Co-design with heterogeneous hardware and software.} User devices are frequently heterogeneous with different hardware and software platforms. As an open question, we can jointly consider different hardware / software platforms and privacy guarantees. For instance, using secure hardware enclaves (e.g., Intel SGX~\cite{costan2016intel} and ARM Trusted Zone~\cite{arm2009security}) to enhance the trustworthy and privacy of federated learning~\cite{moefficient}. 
In co-design with sketches, we can offload the sketching and querying (estimated numbers) parts into the enclave on the user devices, and the aggregation part into the enclave on the server. In this case, we can establish trusted channels between attested user devices and the aggregation server. To further approach level-4 privacy in our taxonomy, we can attempt to privatize the server aggregation logic and the final trained model in secure enclaves. 

\para{Flexible, private federated learning systems.} By enabling sketches in federated learning, we expect to achieve significantly enhanced communication-efficiency and privacy in various distributed learning tasks.  These benefits can encourage more flexible system design choices, e.g., user controllable training time and privacy budget. Users can potentially choose to contribute to large cooperative learning projects without device performance degradation and sensitive information leakage, and decide how accurate and private the transmitted information will be. In addition, our work opens up the possibility to conduct light-weight, private federated learning on low-energy edge devices such as 5G cell towers and mini base stations.

\para{Connection between privacy and communication.} More generally, our work demonstrates that a natural connection exists between privacy preservation and communication reduction in distributed settings. Both tasks share a fundamental aim to reduce, mask, or transform information that is shared across the network in a way that preserves the underlying learning task.  In this work, we capitalize on this connection through the use of sketches. However, it is natural to ask whether other tools from the privacy or distributed learning communities may similarly provide benefits across both dimensions simultaneously.




\bibliographystyle{acm} 
\bibliography{alan,tianrefs}

\end{document}